\documentclass[sigconf]{acmart}
\usepackage{graphicx}
\usepackage{epsfig}
\usepackage{booktabs} 
\usepackage{multirow}
\usepackage{amssymb}
\usepackage{bm}
\usepackage{amsmath}
\usepackage{amsthm}
\usepackage{enumitem}
\usepackage{float}
\usepackage{caption}
\usepackage{subcaption}
\usepackage[export]{adjustbox}
\usepackage[misc]{ifsym}
\usepackage{comment}
\usepackage{soul}
\usepackage{verbatim}
\usepackage{paralist}
\usepackage{xspace}

\usepackage{colortbl}
\definecolor{GAINSBORO}{HTML}{DCDCDC}
\definecolor{LIGHTYELLOW}{HTML}{FFFFE0}
\definecolor{MISTYROSE}{HTML}{FFE4E1}
\definecolor{ALICEBLUE}{HTML}{F0F8FF}

\makeatletter
\newcommand*{\centerfloat}{%
  \parindent \z@
  \leftskip \z@ \@plus 1fil \@minus \textwidth
  \rightskip\leftskip
  \parfillskip \z@skip}
\makeatother

\def \x {\mathbf{x}}
\def \y {\mathbf{y}}

\def \u {\mathbf{u}}
\def \v {\mathbf{v}}

\newcommand{\norm}[1]{\left\lVert #1 \right\rVert}
\newcommand\definefullcoursenames[2]{%
  \expandafter\newcommand\csname var#1var\endcsname{#2}%
}
\newcommand{\fullcoursename}[1]{\csname var#1var\endcsname}
\definefullcoursenames{1}{\text{6.00.1x}}
\definefullcoursenames{2}{\text{6.00.2x}}
\newcommand\defineshortcoursenames[2]{%
  \expandafter\newcommand\csname var#1var\endcsname{#2}%
}
\newcommand{\shortcoursename}[1]{\csname var#1var\endcsname}
\defineshortcoursenames{1A}{\textit{1A}}
\defineshortcoursenames{1B}{\textit{1B}}
\defineshortcoursenames{1C}{\textit{1C}}
\defineshortcoursenames{2A}{\textit{2A}}
\defineshortcoursenames{2B}{\textit{2B}}
\defineshortcoursenames{2C}{\textit{2C}}
\newcommand\definemodelnames[2]{%
  \expandafter\newcommand\csname var#1var\endcsname{#2}%
}
\newcommand{\modelname}[1]{\csname var#1var\endcsname}
\definemodelnames{no-transfer}{\textbf{Label-Truth}}
\definemodelnames{no-transfer-AE}{\textbf{Label-Truth-AE}}
\definemodelnames{naive}{\textbf{Naive Transfer}}
\definemodelnames{in-situ}{\textbf{In-Situ Learning}}
\definemodelnames{instance}{\textbf{Instance-Based Transfer}}
\definemodelnames{passive}{\textbf{Passive-AE Transfer}}
\definemodelnames{active}{\textbf{Active-AE Transfer}}
\makeatletter

\def\th@plain{\thm@preskip\parskip\thm@postskip0pt\itshape}
\def\th@definition{\thm@preskip\parskip\thm@postskip0pt\normalfont}
\def\th@remark{\thm@headfont{\itshape}\normalfont\thm@preskip\parskip\thm@postskip0pt}
\makeatother
\theoremstyle{definition}
\newtheorem{definition}{Definition}[]

\setcopyright{acmlicensed}

\acmDOI{}

\acmISBN{}

\acmConference[LAK'19]{International Learning Analytics and Knowledge Conference}{March 2019}{Tempe, Arizona}
\acmYear{2019}
\copyrightyear{2019}

\acmArticle{}
\acmPrice{}

\editor{}

\begin{document}

\title{Transfer Learning using Representation Learning in Massive Open Online Courses}

\author{Mucong Ding}
\affiliation{%
  \institution{Department of Computer Science and Engineering\\Hong Kong University of Science and Technology\\Hong Kong SAR, China}
}
\email{mcding@connect.ust.hk}

\author{Yanbang Wang}
\affiliation{%
  \institution{Department of Computer Science and Engineering\\Hong Kong University of Science and Technology\\Hong Kong SAR, China}
}
\email{ywangdr@connect.ust.hk}

\author{Erik Hemberg}
\affiliation{%
  \institution{Computer Science and Artificial Intelligence Laboratory\\Massachusetts Institute of Technology\\Cambridge, MA, USA}
}
\email{hembergerik@csail.mit.edu}

\author{Una-May O'Reilly}
\affiliation{%
  \institution{Computer Science and Artificial Intelligence Laboratory\\Massachusetts Institute of Technology\\Cambridge, MA, USA}
}
\email{unamay@csail.mit.edu}

\renewcommand{\shortauthors}{M. Ding et al.}

\begin{abstract}
In a Massive Open Online Course (MOOC), predictive models of student behavior can support multiple aspects of learning, including instructor feedback and timely intervention. Ongoing courses, when the student outcomes are yet unknown, must rely on models trained from the historical data of previously offered courses. It is possible to transfer models, but they often have poor prediction performance. One reason is features that inadequately represent predictive attributes common to both courses. We present an automated transductive transfer learning approach that addresses this issue. It relies on problem-agnostic, temporal organization of the MOOC clickstream data, where, for each student, for multiple courses, a set of specific MOOC event types is expressed for each time unit. It consists of two alternative transfer methods based on representation learning with auto-encoders: a passive approach using transductive principal component analysis and an active approach that uses a correlation alignment loss term. With these methods, we investigate the transferability of dropout prediction across similar and dissimilar MOOCs and compare with known methods. Results show improved model transferability and suggest that the methods are capable of automatically learning a feature representation that expresses common predictive characteristics of MOOCs.
\end{abstract}

%
%



\maketitle

\section{Introduction}
\label{sec:intro}
Massive open online courses (MOOCs) have become popular and provide an inexpensive, learner-directed learning environment. In MOOCs, predictive models of student behavior can support an instructor to improve the student learning, e.g., provide appropriate feedback and timely intervention. Multiple studies have proposed predictive models to, for example, analyze learning progress, support a better understanding of learning abilities, identify at-risk students, and indicate where pro-active interventions may be needed~\cite{chaplot2015predicting, kloft2014predicting}. 

Quantitative, behaviorally driven, predictive models have certain limitations. In particular, while they can be accurate, they may not be interpretable. For example, they do not integrate latent contextual information that is important for elaborating upon the observed behavior such as the fatigue or motivations of a student. Regarding dropout, they do not reveal the students who are not interested in attaining a certificate and others who may be autodidacts who ignore learning design patterns. Additionally, models do not integrate a teacher's goals and intended learning patterns which are complex factors that influence a learner.  Nonetheless, while a course is ongoing and many students are enrolled while out of personal contact with the instructor, automated prediction can be helpful~\cite{baker2014educational}.  An accurate model can identify learners who could benefit from appropriate suggestive interventions which may prevent them from dropping out. It can assist in guiding learners with adaptive instructional materials and pathways to personally appropriate learning resources. Thus, herein, we focus on technical innovation that improves automation and predictive accuracy of models that are useful in transfer learning settings.

There are also technical and practical modeling challenges. Some models predict an outcome that occurs later in the course from the time a student's learning behavior is observed. One example is a model that predicts in week~3 whether a student will dropout in week~4. This type of model is trained, with machine learning, on data retrospectively collected from completed courses. This, however, makes it difficult to use in an ongoing course that differs from the completed course data. Different aspects between courses evolve, including the platform affordances and course design, while the learner cohort also differs.  Despite best efforts, handcrafted features developed for the earlier offering may not express correlations in the ongoing course, i.e., they are brittle. Or, they may not exist in a subsequent offering, i.e., they are infeasible. For example, if a feature is defined on specific course exercises, it cannot be used in a subsequent offering if the exercises are removed. Previous work has relied on handcrafted features for transfer learning and had operational limitations regarding ongoing courses~\cite{boyerTransfer2015}. 

\begin{figure}[t]
    \center
    \includegraphics[width=1.0\linewidth]{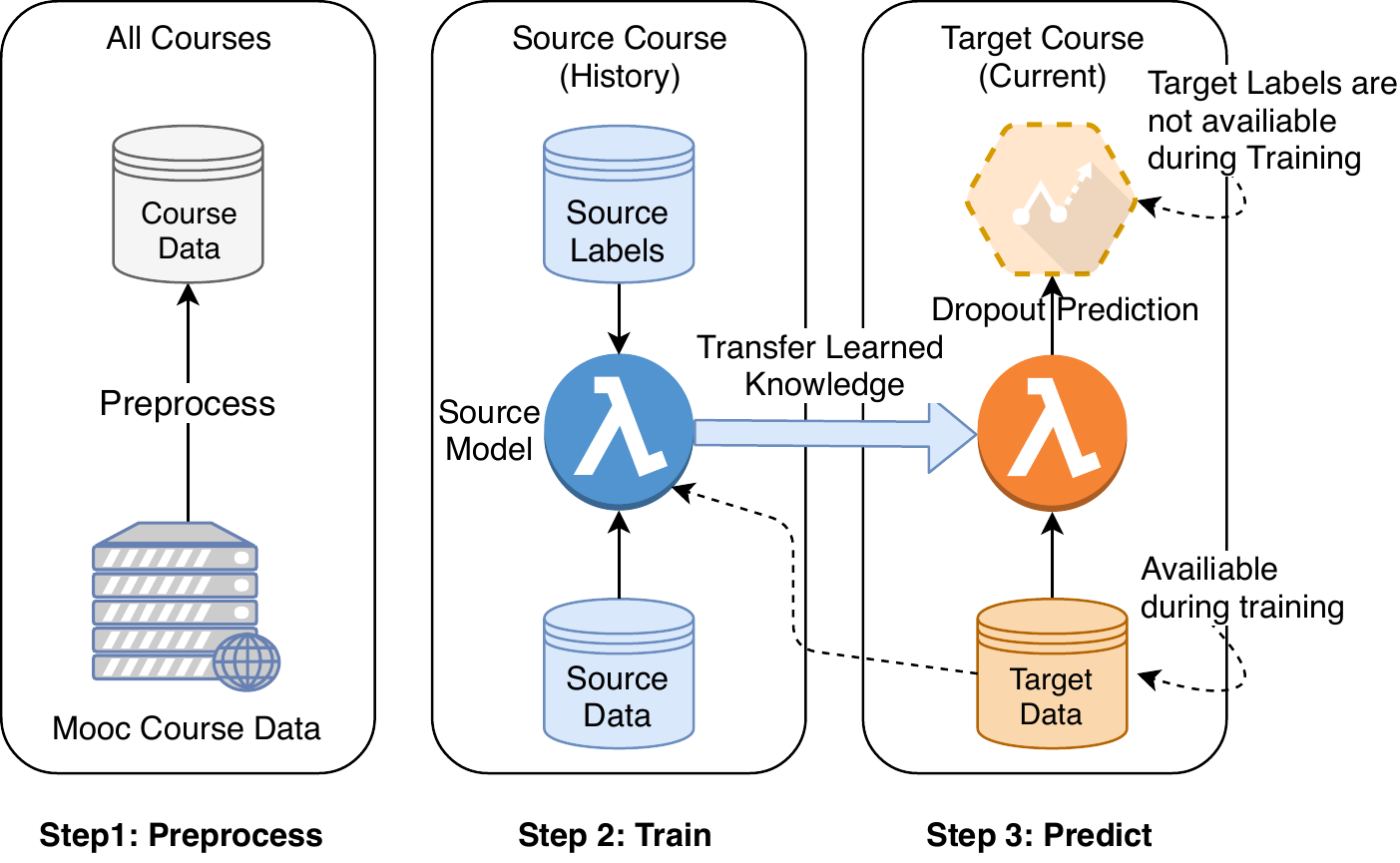}
    \caption{Transfer learning work-flow.}
    \label{fig:transfer_workflow}
\end{figure}

In this paper, our goal is to create operational predictive models for online use. We fundamentally ask whether it is possible, in general, to train a model on a source course's data and reliably transfer it to perform well for an ongoing course. One possible approach is to eliminate customized features engineered by a human that depend on domain knowledge and instead learn a latent representation amenable to the model transfer. Therefore, we propose to investigate transductive transfer learning methods~\cite{pan2010survey}, see step 2 and 3 in Figure~\ref{fig:transfer_workflow}. These methods assume that no label is available for the target task and instead learn a model from the target domain plus a source domain where there are labels and similarity to the target distribution. In a transductive approach known as representation-based transfer learning, a latent space representation is learned automatically using source and target data (the latter without labels). 

Our particular research questions are:
\begin{asparaenum}
\item Does representation learning improve model transfer? We evaluate transferability within offerings for two courses and across two courses.
\item Can representation-based learning work from a universal, basic set of MOOC activity features as input? We test a time-series per student where the frequencies of a set of specific MOOC activity types are expressed per time unit.
\item Can transfer learning improve recognition of minority groups? If we group similar students and transfer learning for each group independently, does predictive performance improve?
\item What are the embedded features that increase the transferability?
\end{asparaenum}

We employ a class of neural networks called auto-encoders (AEs) to compress the input into a latent space representation from which the input is reconstructed, as output. To perform well, the AE has to learn, through the training of its weights, to extract the most relevant features in the representation between the encoder and decoder. For transductive transfer when an AE is trained on both source and target features, its embedding is a set of lower dimensional features that compactly capture mutual properties between the source and target courses, which can then be used by subsequent modeling. For events with strong correlation, e.g. \texttt{play\_video} and \texttt{pause\_video} events, the auto-encoder (AE) learns a compressed representation that reduces the noise and mutual-correlation between them. This leads to improved transferability.  We investigate two variations of representation-based transfer learning:
\begin{inparaenum}
\item \textbf{Post-processing the embedding by transductive principal component analysis}: a passive approach that learns the compact representation before it trains the predictive model.
\item \textbf{Training with correlation alignment loss}: an active approach that trains the auto-encoder and predictive model simultaneously.
\end{inparaenum}

With the learned representation of each method and the predictive model for the target, we can evaluate transferability and compare it with existing methods~\cite{boyerTransfer2015}. See Figure~\ref{fig:transfer_workflow} for the work-flow of our approach. We choose for demonstration dropout prediction in six edX courses -- three offerings of \fullcoursename{1} ``Introduction to Computer Science and Programming Using Python'' and three offerings of \fullcoursename{2} ``Introduction to Computational Thinking and Data Science''.

The paper is organized as follows. In Section~\ref{sec:related_work}, we present related work. In Section~\ref{sec:problem_description} we describe the courses, input data and dropout problem we use for demonstration. This sets up Section~\ref{sec:transfer}, where we present the two transfer learning methods. We evaluate the methods using the courses and dropout prediction problem in Section~\ref{sec:experiments}. In Section~\ref{sec:conclusions} we summarize and mention future work.

\section{Related Work}
\label{sec:related_work}
This section covers related work on transfer learning, representation learning with auto-encoders, feature learning and dropout prediction.

A summary and investigation into transductive transfer learning are provided in ~\cite{pan2010survey,arnold2007comparative}. There are two approaches for transductive feature learning: instance-based methods motivated by importance sampling, and representation-based methods using feature learning. In this paper, we explore representation based transfer methods and use the instance based algorithm as a baseline. A previous study of transfer learning for predictive models in MOOCs uses hand-crafted features and an importance sampling method to shift the source distribution towards the target one~\cite{boyerTransfer2015}. The study also proposes some inductive transfer methods by forcing models to learn from the features within a sliding window, but the resulting transfer methods use the previous week dropout as the target label for the transfer setups and cannot predict the next week dropout well in an online course. A non-MOOC transfer learning study that investigates graduation rates in degree programs~\cite{hunt2017transfer} uses \textit{AdaBoost} and manual features.

Feature identification is a critical precursor to prediction~\cite{gardner2018student}. Some human selected and engineered features are page views, video interactions, forum posts, and content interactions. By human and engineering we imply, respectively, that the choice of the combination is made by learning design experts or researchers and counts of the clickstream elements have to be combined to derive the feature.  An extensive predictive modeling (and coincidentally dropout focused) investigation that relied upon feature engineering at scale on MOOC courses is ~\cite{veeramachaneni2014towards}. The same study made use of crowdsourcing to engineer features.  Herein we forgo feature engineering for feature learning, in the context of transfer learning. We are preceded by a study for MOOCs that used predictive engagement analytics with long short-term memory (LSTM) networks for feature learning~\cite{lecommunication}.

Transfer learning is applicable to predicting any outcome variable, and we demonstrate it with dropout prediction. There are many types of predictive models for MOOCs, e.g., dropout, certification and grade~\cite{gardner2018student,gardner2018morf}. There are a variety of related papers which investigate the dropout prediction problem~\cite{chaplot2015predicting, kloft2014predicting, burgos2017data, brooks2015timeb}. Most existing work use machine learning approaches such as logistic regression (LR), support vector machine (SVM), decision trees (DT) and Neural Networks (NN).

\section{Problem Description}
\label{sec:problem_description}
In this section, we compare and contrast the courses we use to evaluate the representation-based transfer learning methods we purposed. We describe how we organize student activity data collected from them for input to the methods. Finally, we define the predictive modeling problem, dropout, that we use to demonstrate the transfer learning methods on.

\subsection{Input Data Organization}
In practical online prediction, some student attributes are unavailable (e.g., certificate and registration information) and cannot be used to filter the set of students. Thus, the data used for prediction exhibits high variance. We use the click-stream log events, which were engineered by the learning platform developers without a prediction problem in mind, as input to the transfer learning methods. The list of events can be found in the legend of Figure~\ref{fig:event_frequencies}. All events have a \textit{time-stamp} and \textit{event type}. Therefore it is straightforward to aggregate the events by week and event type, per student. This results in a multivariate time series, for each student, $[\x_1, \x_2, \cdots, \x_T]$, where each $\x_k$ is a vector of the normalized frequencies of the event types in that week $k$, and $T=9$ is the number of weeks. The sets of event types of courses do not need to be identical. However, the most frequent and important event types often overlap. 

\subsection{Courses}
We experiment with two courses offered on the edX MOOC platform: \textit{Introduction to Computer Science and Programming Using Python} (\fullcoursename{1}), and \textit{Introduction to Computational Thinking and Data Science} (\fullcoursename{2}). We have 3 offerings of each course.

In terms of structure, all offerings have 9 weeks. Both courses have multiple units, where each unit has an associated graded problem set. Students are expected to watch lecture videos narrated by instructors and complete ``finger exercises'' - optional problems interspersed in lecture videos that teach the content discussed in the video. The topics of each course differ because one course is the continuation of the other, see Table~\ref{tab:graded_activities} for details. The quantities of videos and finger exercises are much higher in 6.00.1x, see Table~\ref{tab:number_of_items}. Enrollment and activity volume measured in click-stream events are shown in Table~\ref{tab:course_statistics}. 

\begin{table}[t]
\centerfloat
\begin{tabular}{c|l|c}
\textbf{Course}                   & \textbf{Assignment}         & \textbf{Due Week} \\ \hline
\multirow{5}{*}{\textbf{6.00.1x}} & Python Basics               & Week 4            \\
                                  & Simple Programs             & Week 5            \\
                                  & Structured Types            & Week 7            \\
                                  & Good Programming Practices  & Week 8            \\
                                  & Object Oriented Programming & Week 9            \\ \hline
\multirow{5}{*}{\textbf{6.00.2x}} & Optimization                & Week 5            \\
                                  & Randomness                  & Week 6            \\
                                  & Midterm Exam                & Week 7            \\
                                  & Statistics                  & Week 8            \\
                                  & Modeling and Fit            & Week 9            \\ \hline
\end{tabular}
\caption{6.00.1x and 6.00.2x Assignments.}
\label{tab:graded_activities}
\end{table}

\begin{table}[t]
\centerfloat
\begin{tabular}{c|c|c}
\textbf{Course}  & \textbf{N. Videos} & \textbf{N. Finger Exercises} \\ \hline
\textbf{6.00.1x} & 81                 & 555                          \\
\textbf{6.00.2x} & 43                 & 177                          \\ \hline
\end{tabular}
\caption{Resource quantities in terms of video and finger exercises. 6.00.1x has more resources than 6.00.2x.}
\label{tab:number_of_items}
\end{table}

\begin{table*}[t]
\centerfloat
\resizebox{\linewidth}{!}{%
\begin{tabular}{c|c|c|c|c|c|c}
\textbf{Course}      & \multicolumn{3}{c|}{\textbf{6.00.1x}}                                            & \multicolumn{3}{c}{\textbf{6.00.2x}}                                        \\ \hline
\textbf{Offering}    & Summer 2016A(\shortcoursename{1A}) & Summer 2016B(\shortcoursename{1B}) & Spring 2017(\shortcoursename{1C}) & Spring 2016(\shortcoursename{2A}) & Fall 2016(\shortcoursename{2B}) & Spring 2017(\shortcoursename{2C}) \\ \hline
\textbf{N. students} & 37,363                    & 15,199                    & 26,011                   & 6,774                    & 6,945                  & 5,893                    \\ \hline
\textbf{N. events}   & 17,333,974                & 7,900,908                 & 13,176,220               & 2,642,528                & 2,501,276              & 2,034,539                \\ \hline
\end{tabular}
}
\caption{Summary of the course statistics regarding number of students and events for \fullcoursename{1} and \fullcoursename{2}. We use the symbols in brackets to note the six courses through the rest of the paper.}
\label{tab:course_statistics}
\end{table*}

\begin{figure}[t]
    \centerfloat
    \includegraphics[width=1\linewidth]{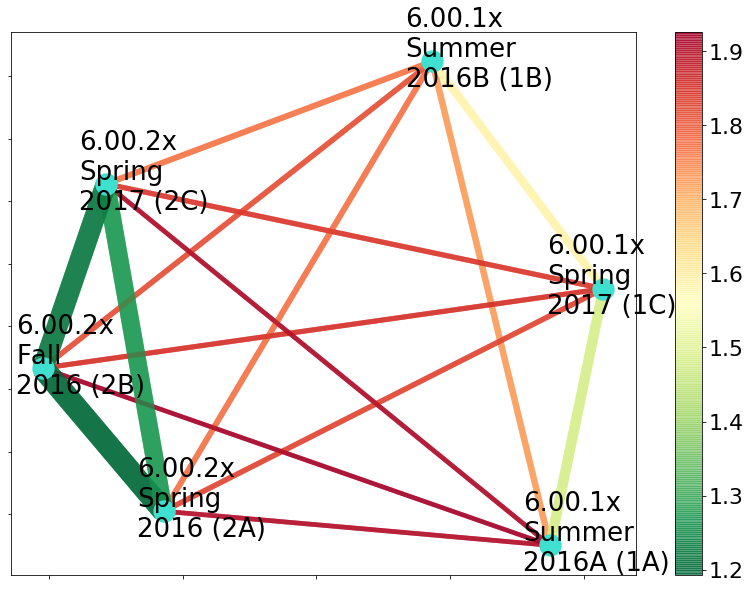}
    \caption{2D-embedding by MDS of the Proxy A-distances (PADs) between features from different courses. The colorbar shows the distance value. The color and thickness of a line segment are proportional to the pairwise distance (Note, the range of PAD is defined in $[0, 2]$).}
    \label{fig:dissimilarity_embedding}
\end{figure}

We can statistically compare the courses to gain a mathematical estimation of similarity.  We use the Proxy A-distance (an approximation of the $\mathcal{H}$-divergence~\cite{ben2010theory}) as an indicator of the similarity between two samples where each sample distribution consists of the frequency of events with an arbitrary type that occurred in an arbitrary week of a specific course. The pair-wise Proxy A-distances (PADs) are difficult to visualize, so we first compute a 2D-embedding where the distances between different distributions of different courses are the average PADs. We find the 2D-embedding which best preserves the distances by multi-dimensional scaling. We visualize the pairwise distances in Figure~\ref{fig:dissimilarity_embedding}. This identifies 6.00.2x Spring 2016 (\shortcoursename{2A}) and 6.00.2x Fall 2016 (\shortcoursename{2B}) as most similar while 6.00.1x Summer 2016A (\shortcoursename{1A}) and 6.00.2x Fall 2016 (\shortcoursename{2B}) are most dissimilar. This is mainly because \shortcoursename{2A} and \shortcoursename{2B} are two offerings of the same course, while \shortcoursename{1A} is a different course.

\subsection{Dropout Prediction}\label{sec:dropout}
\begin{figure}[t]
    \centerfloat
    \includegraphics[width=1\linewidth]{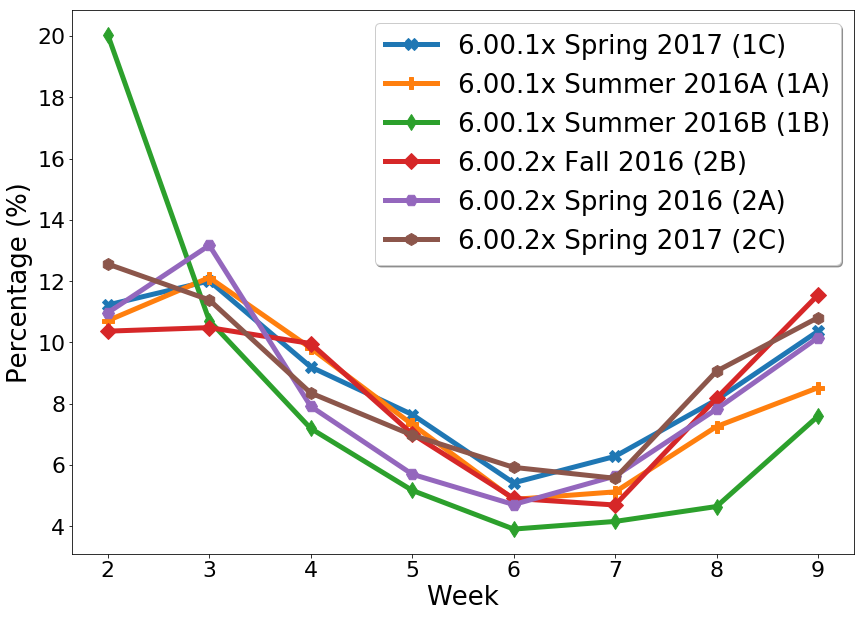}
    \caption{Percentage of students dropout in each week. X-axis shows the week and Y-axis the percentage of students that dropped out, the color shows the course.}
    \label{fig:dropout_percentages}
\end{figure}

We adopt a widely used dropout definition: dropout occurs when the student no longer interacts with the MOOC platform. When considering all click-stream event types, the dropout labels become noisy. Therefore we define dropout based on video events (e.g., \texttt{play\_video}). For a time-granularity based on weeks, the dropout week of a student is defined as the week after the student's last video interaction event. By this definition, students cannot drop out in the first week, and we train one predictive model for each week after the first. The percentages of student dropout in each week are shown in Figure~\ref{fig:dropout_percentages}.

Prediction can be based on data from the first week up to the week before the prediction. A student is characterized by a pair of time-series $(\x, \y)$. The label of a student is a uni-variate time-series $\y = [y_1, \cdots, y_T]$, where $y_k \in \{\textnormal{False}, \textnormal{True}\}$ indicates whether the student has dropped-out in week $k$ and $T$ is the total number of weeks. The features of a student, can be represented as a multi-variate time series $\x = [\x_1, \cdots, \x_T]$, where $x_k$ is a set of features in week $k$. To predict student dropout during the course, we train $T-1$ models $[f_2(\cdot), \cdots, f_T(\cdot)]$, one for each week, where $f_k(x_k)$ is the prediction for week $k$. The model $f_k(\cdot)$ for week $k$ uses $y_k$ as label, and $[\x_1, \cdots, \x_{k-1}]$ as features. Each model solves a binary classification problem, i.e. did the student stay in the course or dropout.

\section{Transfer Learning}
\label{sec:transfer}
In this section, we formulate the problem of transferring models between MOOC courses and then introduce our transfer methods.

\subsection{Transfer Learning Definition}
\label{subsec:transfer_definition}
A domain $\mathcal{D}$ consists of two components: a feature space $\mathcal{X}$ and a marginal probability distribution $\Pr(X)$ where $X = \{x_i\}_{i=1}^n\subset\mathcal{X}$. Given a specific domain, a task $\mathcal{T}$ consists of two components: a label space $\mathcal{Y}$ and a conditional probability distribution $\Pr(Y|X)$ where $Y = \{y_i\}_{i=1}^n\subset\mathcal{Y}$. Considering a source course $S$ and target course $T$, the feature spaces of the source and target domains $\mathcal{D}_S$ and $\mathcal{D}_T$ are the same but the feature distributions are different $\Pr(X_S)\neq\Pr(X_T)$. However, the prediction tasks $\mathcal{T}_S$ and $\mathcal{T}_T$ for the two domains are the same, as the conditional distributions coincide $\Pr(X_S|Y_S)=\Pr(X_T|Y_T)$.

\begin{definition}
\textbf{Transductive Transfer Learning Problem in MOOCs}
In transfer learning the training of the target predictive function $f_T(\cdot)$ in the target domain $\mathcal{D}_T$ is supplemented using the knowledge in the source domain $\mathcal{D}_S$, where $\mathcal{D}_S\neq\mathcal{D}_T$ and $\mathcal{T}_S=\mathcal{T}_T$. At training time, source data $\{x_{S_i}, y_{S_i}\}_{i=1}^{n_S}$ and the unlabeled target domain features $X_T=\{x_{T_i}\}_{i=1}^{n_T}$ are available. 
\end{definition}

\subsection{Transfer Learning Methods}
\label{subsec:transfer_methods}
We use auto-encoders to learn a representation space that is common to both source and target domains. By training an auto-encoder (AE) on both of the source and target features (and, in some variants, the source labels) and using a learning signal that measures the output's distance from the input, the AE's embedding layer between the encoder and decoder is forced to capture the common characteristics of the two distributions. There is a trade-off between the prediction model's capability and the dimensionality of the representation space. When the reduction is small, the auto-encoder can learn multiple modes in the distribution but the predictor is prone to overfit to the source task. When the reduction is large, the auto-encoder can only learn a single mode presenting the risk that the embedding is not predictive enough. Thus a dimensionality should be carefully chosen for a specific combination of model and data set.

We introduce two different transfer learning methods: \begin{inparaenum}[\itshape 1)]
\item Passive Transfer with Transductive-PCA (\modelname{passive})
\item Active Transfer with CORAL loss (\modelname{active})
\end{inparaenum}

\subsubsection{Passive Transfer with Transductive PCA (T-PCA)}
\begin{figure*}[t]
    \centerfloat
    \includegraphics[width=0.7\linewidth]{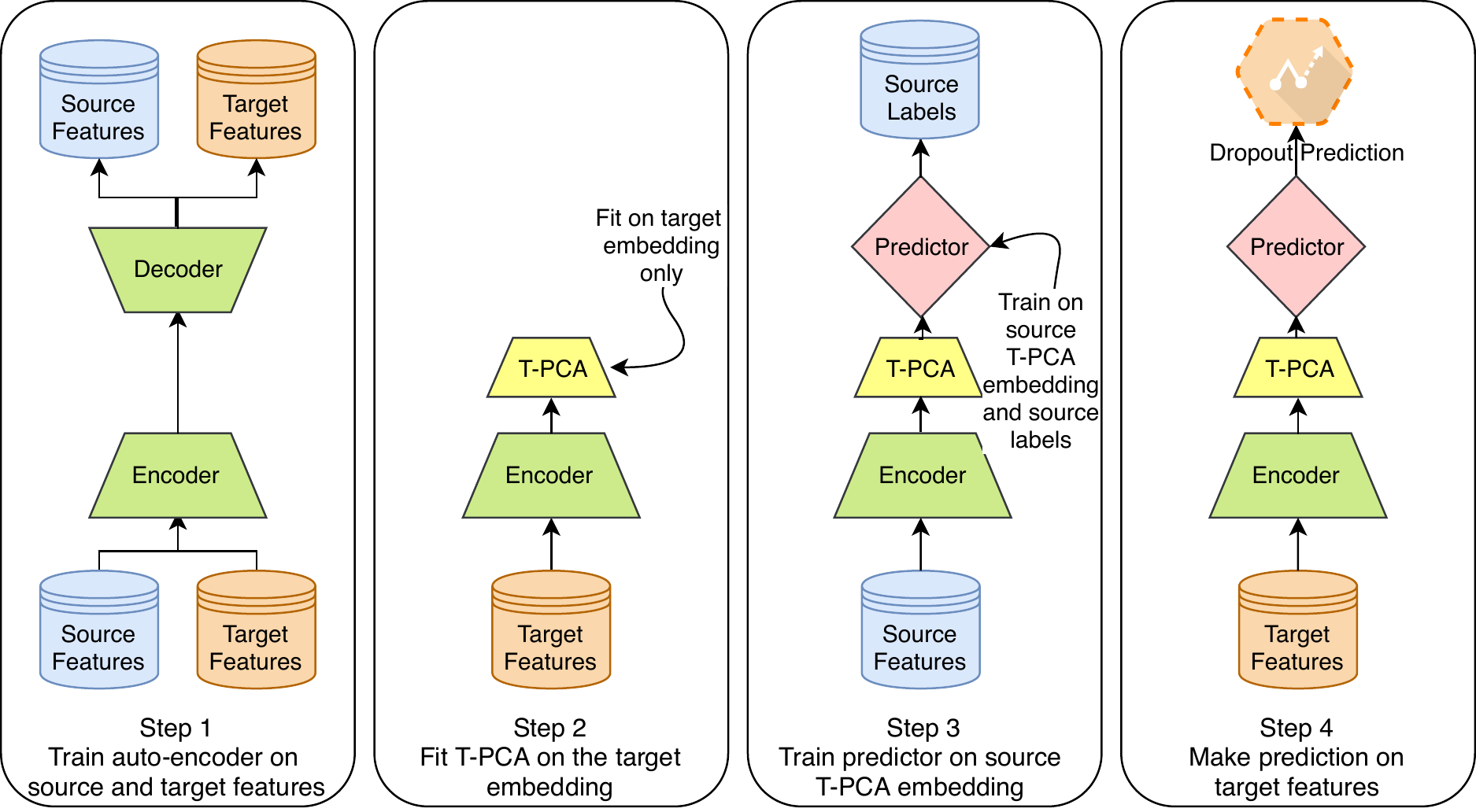}
    \caption{Passive transfer with transductive PCA.}
    \label{fig:passive_transfer_schematic}
\end{figure*}

The work-flow is depicted in Figure~\ref{fig:passive_transfer_schematic}. The target embedding is obtained in a ``passive'' way, since no objective functions are defined and used. First, an AE is trained on both the source and target features. Next, (Step 2 in Figure~\ref{fig:passive_transfer_schematic}), using target features only, a PCA~\cite{jolliffe1986principal} transform is fit on the learned target embedding to transform the source embedding for predictive model training (Step 3)~\cite{mesnil2011unsupervised}. This avoids the prediction model being trained on an embedding that has learned irrelevant features from the source domain. The number of outputs of T-PCA is set to be larger than the number of predicted labels.

\subsubsection{Active Transfer with CORAL loss}
\begin{figure}[t]
    \centerfloat
    \includegraphics[width=\linewidth]{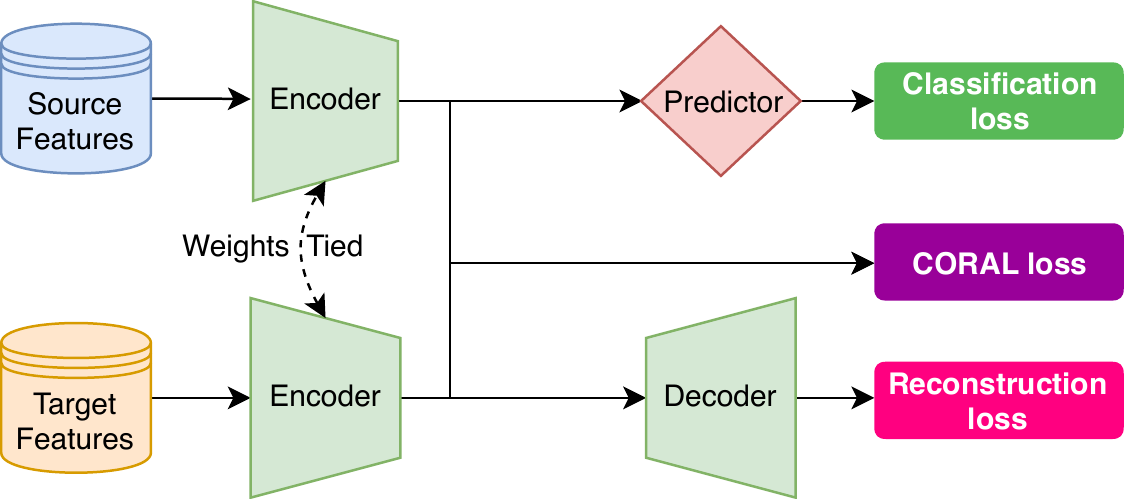}
    \caption{Active transfer with CORAL loss.}
    \label{fig:active_transfer_schematic}
\end{figure}

An unsupervised transfer learning method, CORAL performs a transformation to align the second-order statistics of the source and target domains~\cite{sun2016return}. This removes variations that are only present in the source domain in the learned embedding. A general transform is achieved a by deep neural network~\cite{sun2016deep}. A CORAL loss is introduced as a term in the objective function because minimizing the prediction loss itself can lead to overfitting to the source domain, causing reduced performance on the target domain. Let $C_S$ and $C_T$ indicate the covariance matrices of the source and target embedding $E_S=\{\u_i\}_{i=1}^{n_S}$ and $E_T=\{\v_i\}_{i=1}^{n_T}$, the CORAL loss~\cite{sun2016deep} is defined as $\mathcal{L}_{CORAL} = \frac{1}{4d^2}\norm{C_S-C_T}_F^2$ where $\norm{\cdot}_F$ denotes the Frobenius norm, and $d$ is the dimension of the common embedding space $\u, \v\in\mathcal{E}\subseteq\mathbb{R}^d$. While minimizing the CORAL loss alone can lead to degenerated features, jointly training the auto-encoder and the embedding predictor with both losses and also the target reconstruction loss tries to strike a balance (Figure~\ref{fig:active_transfer_schematic}).

\subsection{Auto-Encoders and Predictive Models}
Both transfer learning methods use an AE and a predictive model, and for both autoencoding and prediction, neural networks can be used. In preliminary AE studies, we used a neural network implementation of PCA as an AE baseline. We then explored deep networks: a multi-layer perceptron, a convolutional neural network (CNN) (the order of course materials in each week preserves the implicit dependence in the knowledge graph, so we can apply 1D-convolution in the time domain to find the implicit learning patterns) and a Long Short-Term Memory (LSTM) network adapted for autoencoding. The LSTM AE has two 1D-convolutional layers with kernel size 1 and a bi-directional LSTM layer in the encoder and one 1D-convolutional layer with kernel size 1 and a bi-directional LSTM layer in the decoder (see Appendix~\ref{appendix:model_architectures} for details). In a bi-directional LSTM, two hidden layers of opposite directions are connected to the same output to make future input information reachable from the current state. We report our best results which use the LSTM AE.

Training models for prediction relies on the embedded features from the AE (or T-PCA) representations. We explored predictive modeling both with and without embedding features. We used logistic regression (LR) as a baseline of the predictive models and then compared using CNN and LSTM with architectures similar to the encoder parts of AEs. We report our best results which use CNN on top of embeddings and LSTM without embeddings (see Appendix~\ref{appendix:model_architectures} for details).

\section{Experiments}
\label{sec:experiments}
We performed a massive number of experiments, which was made possible by using an efficient and scalable software framework that allows model tuning. Preliminary experiments allowed us to settle on parameters, e.g., architectures of the networks and training hyper-parameters to use throughout evaluations. We ran our entire pipeline multiple times with different combinations of models and architectures. The pipeline is released in an open-source \texttt{Python} MOOC learner data science analysis (\texttt{mldsa}) toolkit \footnote{\url{https://github.com/MOOC-Learner-Project/MOOC-Learner-Data-Science-Analytics}}.

We use the area under the receiver operating characteristic curve (AUC) to measure the performance of all predictive models. The AUC scores are chosen from the best performing architecture and are averaged on all possible pairs of source and target courses and on all weeks. See Appendix~\ref{appendix:experimental_configurations} for detailed experimental configurations.

\begin{figure}[t]
    \centerfloat
    \begin{subfigure}[b]{0.8\linewidth}
		\includegraphics[width=\linewidth]{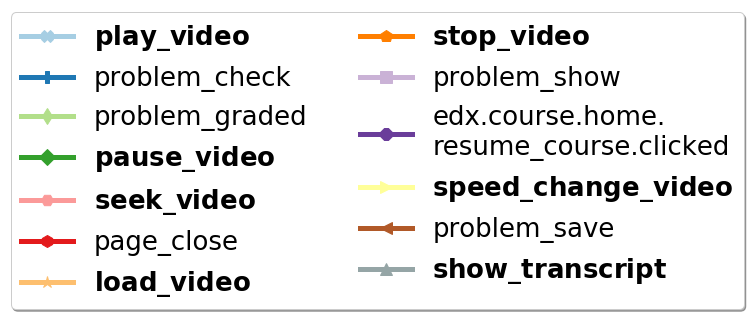}
	\end{subfigure}
	
	\begin{subfigure}[b]{1.1\linewidth}
		\centerfloat
		\includegraphics[width=\linewidth]{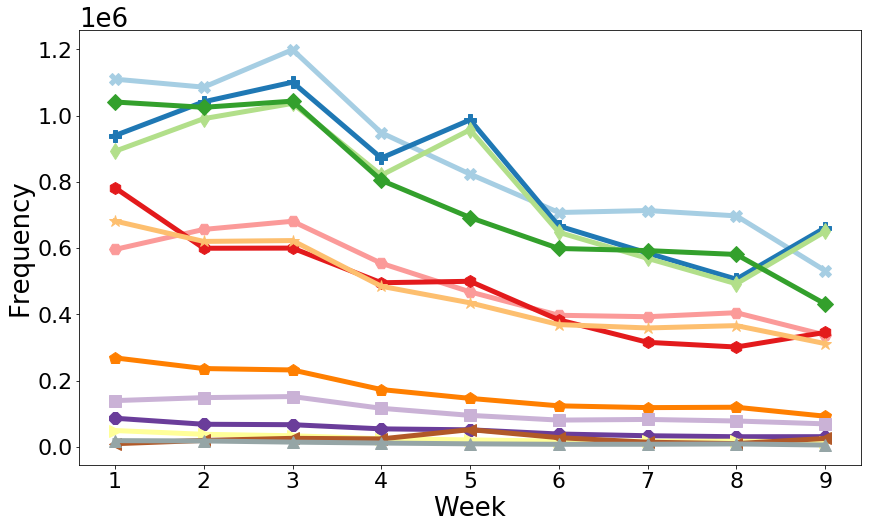}
	\end{subfigure}
    \caption{The total frequencies of the thirteen common event types used to define the raw features in all the six courses. X-axis shows the week and Y-axis the frequency, the color shows the event type. Bold indicates video event. The events tend to decrease and some are correlated.}
    \label{fig:event_frequencies}
\end{figure}

Our input features are based on the thirteen common event types, see Figure~\ref{fig:event_frequencies}. These events all tend to decrease in frequency, most likely due to dropout. We also note that some events are highly correlated (e.g. \texttt{problem\_check} and \texttt{problem\_graded}). This is one of the motivations of representation learning on the raw features.

\subsection{Transfer Learning Baselines}
We set up four baseline models. All of them use LSTM or LSTM AE and CNN for prediction (see Appendix~\ref{appendix:model_architectures}), the same as \modelname{passive} and \modelname{active}.
\begin{asparadesc}
    \item[\modelname{no-transfer} and \modelname{no-transfer-AE}] These models do not use transfer learning. Instead, they are trained with target features and labels (retrieved retrospectively) and provide a label truth baseline. \modelname{no-transfer-AE} uses an LSTM AE embedding for prediction. The target domain data is split into train and test, and test accuracy is reported. 
    \item[\modelname{naive}] This model is learned on the source course (without using target labels) and subsequently applied to the target course. 
    \item[\modelname{in-situ}] We learn a predictive model from the data from the on-going course itself with a sliding window~\cite{boyerTransfer2015}. For week $k\leq3$, we use window size $w=k-2$ and apply the model trained for the previous week $k-1$ for prediction. It transfers between weeks of the same target course and does not use the source course.
    \item[\modelname{instance}] Here it is assumed the features of the source and target domains are drawn from a common distribution and the difference comes from a sample selection bias. We correct this bias by giving more weight to the students in the source course that are similar to the students in the target course. The learning objective in the target domain $\Sigma_{i=1}^{n_T}l(x_{T_i}, y_{T_i}, \theta)$ is approximated by $\Sigma_{i=1}^{n_S}\Pr(x_{T_i})/\Pr(x_{S_i})l(x_{S_i}, y_{S_i}, \theta)$. Thus, by applying different loss weights $\Pr(x_{T_i})/\Pr(x_{S_i})$ to each instance $x_{S_i}$ in the source domain we can train a precise model for the target domain. We use the kernel-mean matching algorithm~\cite{huang2007correcting} to compute the weights.
\end{asparadesc}

\subsection{Model Selection}
\begin{figure}[t]
	\centerfloat
	\includegraphics[width=1.05\linewidth]{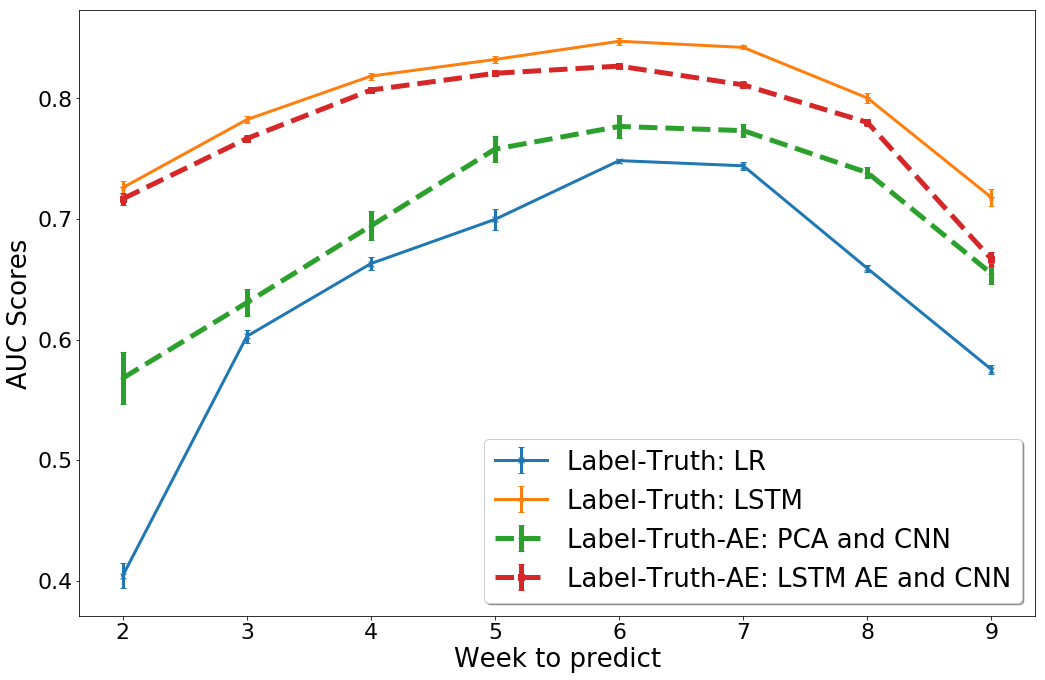}
	\caption{Average AUC scores (with error bars) of predictors of different architectures on the six courses over different weeks. We can see LSTM consistently outperforms LR, and the prediction performance of CNN on embedding learned by LSTM AE (which consists of eight features per week) is close to the best performance of predictors on raw features.}
	\label{fig:hypothesis_predictor_performance}
\end{figure}

First, without transfer, we compared the prediction performance of the LSTM neural network with the standard logistic regression (LR), see Figure~\ref{fig:hypothesis_predictor_performance}. The LSTM model consistently outperforms LR by a large margin. Thus we chose LSTMs.

Next, we compared the performance of predictive models using the embeddings from the AEs as input. More specifically, a CNN model trained on an embedding learned by an LSTM AE or PCA, see Figure~\ref{fig:hypothesis_predictor_performance}. With bottleneck size of eight, the AUC scores of the CNN and LSTM AE embedding are similar to the LSTM's. This implies that it is possible to learn a compact and effective embedding, and suggests the effective use of dimensionality reduction for transfer learning. All predictive models perform better in the middle part of the course, and their AUCs are negatively correlated with the dropout rates (see Figure~\ref{fig:dropout_percentages}).

\subsection{Transfer Learning Results}

\begin{table*}[t]
    \centerfloat
    \begin{tabular}{c|c|c|c|c|c|c|c}
    \rowcolor{GAINSBORO}
                                  & \shortcoursename{1B}$\rightarrow$\shortcoursename{1A}   & \shortcoursename{1C}$\rightarrow$\shortcoursename{1A}   & \shortcoursename{1A}$\rightarrow$\shortcoursename{1B}   & \shortcoursename{1C}$\rightarrow$\shortcoursename{1B}   & \shortcoursename{1A}$\rightarrow$\shortcoursename{1C}   & \shortcoursename{1B}$\rightarrow$\shortcoursename{1C}   & Avg.       \\ \hline
    \rowcolor{LIGHTYELLOW}
    \modelname{passive}      & .758$\pm$1          & .777$\pm$1          & .749$\pm$2          & .757$\pm$2          & .801$\pm$1          & .784$\pm$2          & .771$\pm$2 \\ \hline
    \rowcolor{LIGHTYELLOW}
    \modelname{active}       & \textbf{.759$\pm$1} & \textbf{.788$\pm$1} & \textbf{.769$\pm$2} & \textbf{.769$\pm$2} & \textbf{.812$\pm$1} & \textbf{.792$\pm$2} & \textbf{.782$\pm$1} \\ \hline \hline
    \rowcolor{MISTYROSE}
    \modelname{in-situ}      & \multicolumn{2}{c|}{.681$\pm$2}           & \multicolumn{2}{c|}{.659$\pm$2}           & \multicolumn{2}{c|}{.724$\pm$2}           & .688$\pm$2 \\ \hline
    \rowcolor{MISTYROSE}
    \modelname{instance}     & .613$\pm$1          & .683$\pm$1          & .608$\pm$1          & .663$\pm$1          & .653$\pm$1          & .649$\pm$1          & .645$\pm$1 \\ \hline
    \rowcolor{MISTYROSE}
    \modelname{naive}        & .716$\pm$1          & .756$\pm$1          & .700$\pm$3          & .699$\pm$3          & .743$\pm$2          & .736$\pm$2          & .725$\pm$2 \\ \hline \hline
    \rowcolor{ALICEBLUE}
    \modelname{no-transfer}     & \multicolumn{2}{c|}{.800$\pm$1}            & \multicolumn{2}{c|}{.773$\pm$1}           & \multicolumn{2}{c|}{.819$\pm$1}           & .797$\pm$1 \\ \hline
    \end{tabular}
    \caption{\fullcoursename{1}$\rightarrow$\fullcoursename{1}: Average AUC scores of transfer methods and the Label-Truth (no-transfer) baseline for all weeks. Transfer between offerings of the same course.}
    \label{tab:gorup11_model_comparison}
\end{table*}

\begin{table*}[t]
    \centerfloat
    \begin{tabular}{c|c|c|c|c|c|c|c|c|c|c}
    \rowcolor{GAINSBORO}
                                  & \shortcoursename{2A}$\rightarrow$\shortcoursename{1A} & \shortcoursename{2B}$\rightarrow$\shortcoursename{1A} & \shortcoursename{2C}$\rightarrow$\shortcoursename{1A} & \shortcoursename{2A}$\rightarrow$\shortcoursename{1B} & \shortcoursename{2B}$\rightarrow$\shortcoursename{1B} & \shortcoursename{2C}$\rightarrow$\shortcoursename{1B} & \shortcoursename{2A}$\rightarrow$\shortcoursename{1C} & \shortcoursename{2B}$\rightarrow$\shortcoursename{1C} & \shortcoursename{2C}$\rightarrow$\shortcoursename{1C} & Avg.       \\ \hline
    \rowcolor{LIGHTYELLOW}
    \modelname{passive}      &\textbf{.753$\pm$1} &\textbf{.752$\pm$1} &\textbf{.760$\pm$1} &\textbf{.735$\pm$2} & .733$\pm$2         & \textbf{.745$\pm$2} &\textbf{.786$\pm$1} &\textbf{.777$\pm$1} &\textbf{.782$\pm$1}   &\textbf{.758$\pm$1}\\ \hline
    \rowcolor{LIGHTYELLOW}
    \modelname{active}       & .713$\pm$1         & .720$\pm$1         & .717$\pm$1         & .734$\pm$2         &\textbf{.740$\pm$2} & .743$\pm$2          & .755$\pm$1         & .751$\pm$2                & .757$\pm$2           & .737$\pm$2 \\ \hline \hline
    \rowcolor{MISTYROSE}
    \modelname{in-situ}      & \multicolumn{3}{c|}{.681$\pm$2}                              & \multicolumn{3}{c|}{.659$\pm$2}                               & \multicolumn{3}{c|}{.724$\pm$2}                                       & .688$\pm$1 \\ \hline
    \rowcolor{MISTYROSE}
    \modelname{instance}     & .590$\pm$2         & .569$\pm$2         & .606$\pm$1         & .637$\pm$3         & .640$\pm$2         & .600$\pm$1          & .579$\pm$2         & .528$\pm$1                & .598$\pm$3           & .594$\pm$2 \\ \hline
    \rowcolor{MISTYROSE}
    \modelname{naive}        & .668$\pm$1         & .700$\pm$2         & .706$\pm$2         & .675$\pm$3         & .684$\pm$3         & .676$\pm$3          & .735$\pm$2         & .716$\pm$2                & .739$\pm$2           & .702$\pm$2 \\ \hline \hline
    \rowcolor{ALICEBLUE}
    \modelname{no-transfer}     & \multicolumn{3}{c|}{.800$\pm$1}                              & \multicolumn{3}{c|}{.773$\pm$1}                               & \multicolumn{3}{c|}{.819$\pm$1}                                       & .797$\pm$1 \\ \hline
     \end{tabular}
    \caption{\fullcoursename{2}$\rightarrow$\fullcoursename{1}: Average AUC scores of transfer methods and the Label-Truth (no-transfer) baseline on all weeks. Transfer between courses.}
    \label{tab:gorup21_model_comparison}
\end{table*}

\begin{figure*}[t]
	\centerfloat
	\begin{subfigure}[c]{0.42\linewidth}
		\centerfloat
		\includegraphics[width=\linewidth]{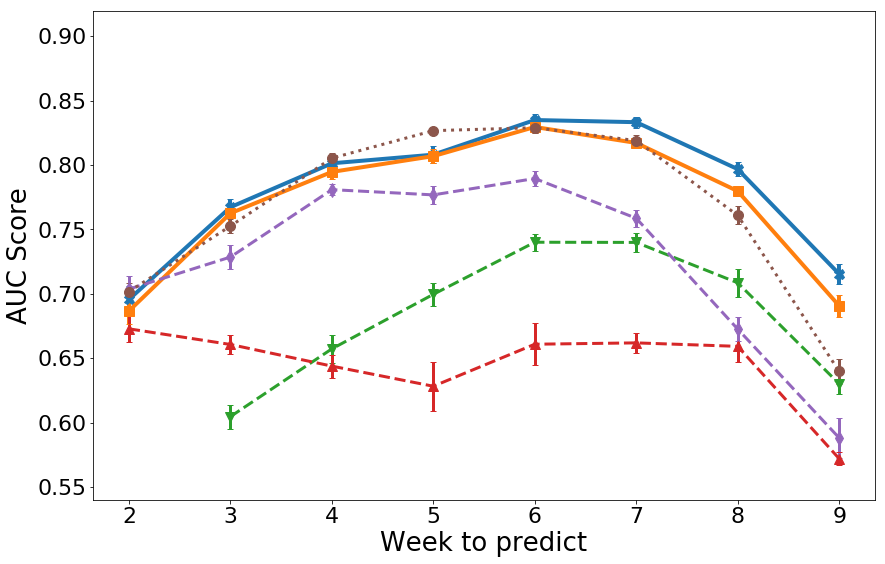}
		\caption{\label{fig:transfer_performances_1} Average AUCs on the six combinations of\\similar courses  (\fullcoursename{1}$\rightarrow$\fullcoursename{1}).}
	\end{subfigure}
	\begin{subfigure}[c]{0.42\linewidth}
		\centerfloat
		\includegraphics[width=\linewidth]{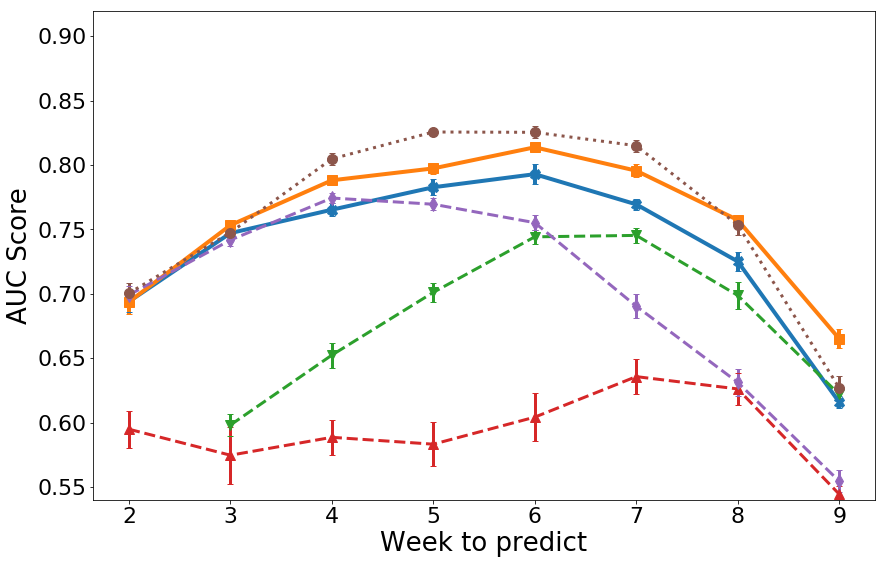}
		\caption{\label{fig:transfer_performances_2} Average AUCs on the nine combinations of\\dissimilar courses  (\fullcoursename{2}$\rightarrow$\fullcoursename{1}).}
	\end{subfigure}
	\begin{subfigure}[c]{0.16\linewidth}
		\centerfloat
		\includegraphics[width=\linewidth]{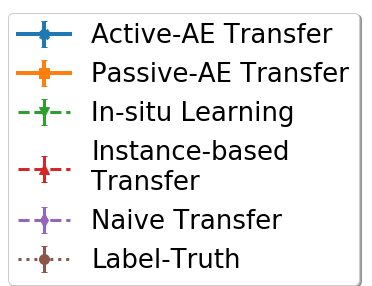}
	\end{subfigure}
	\caption{Average AUC scores (with error bars) of transfer methods and the Label-Truth (no-transfer) baseline for each week on different groups of source and target combinations.}
	\label{fig:transfer_performances}
\end{figure*}

We now compare the proposed transfer learning methods with the baselines and \modelname{no-transfer} case. The transfer methods that perform best are the \modelname{passive} and \modelname{active}, even similar to the \modelname{no-transfer} baseline. The dropout prediction performance of \modelname{passive} and \modelname{active} transfer with the baselines (\modelname{no-transfer}, \modelname{no-transfer-AE}, \modelname{naive}, \modelname{in-situ}, and \modelname{instance}), for the similar pairs of source and target \fullcoursename{1}$\rightarrow$\fullcoursename{1} (within offerings of one course) are shown in Table~\ref{tab:gorup11_model_comparison} and Figure~\ref{fig:transfer_performances_1}, and the dissimilar pairs \fullcoursename{2} $\rightarrow$ \fullcoursename{1} (across two courses) in Table~\ref{tab:gorup21_model_comparison} and Figure~\ref{fig:transfer_performances_2}. Note that the performance of \modelname{in-situ} and \modelname{no-transfer} does not depends on the choice of source course. All transfer baselines have significantly lower average AUCs than the \modelname{passive} and \modelname{active} methods. 

Now we analyze the average AUC per predicted week, see Figure~\ref{fig:transfer_performances}. It shows that \modelname{naive} overfits to the source domain from week 5. \modelname{in-situ} learning does not work well until the final weeks of the course since it applies a sliding window and a considerable proportion of information is lost when week $k$ is small. \modelname{instance} has abysmal performance on all weeks, since the raw features we used are high-dimensional and sparsely distributed, and hence obtaining a good approximation of the sampling weights is very challenging. We note that \modelname{in-situ} and \modelname{instance} struggles to outperform \modelname{naive} in most cases. An important reason for this is that we used a very low-level input representation (time-series of click-stream events). These methods need complex features whereas representation learning is capable of learning from simpler ones. We anticipate an improved performance for the baseline transfer methods with handcrafted features, but that is not certain and also demands a human to solve the challenge of engineering good features. 

\begin{figure}[t]
	\centerfloat
	\includegraphics[width=1\linewidth]{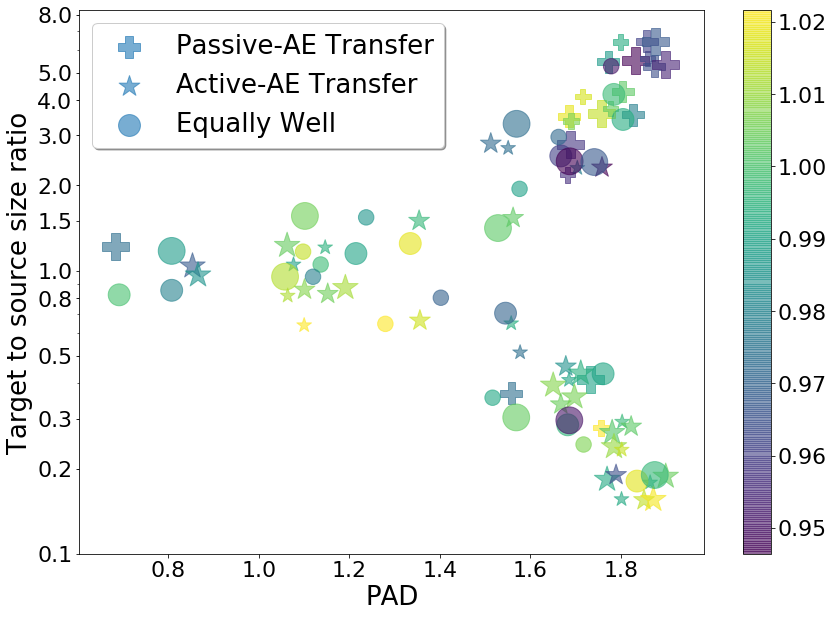}
	\caption{Scatter plot of all \modelname{passive} and \modelname{active} experiments. X-axis shows the Proxy A-distances (PAD) between the source and target input features (sub-sequences till week $k-1$ when predicting for week $k$). Y-axis shows the ratios between the number of students till week $k-1$ in the target and source course. The shape shows the wining method among \modelname{passive} and \modelname{active}. If the difference between their AUCs is less than $1\%$ of their averages, we consider they perform equally well. The color shows the ratio between its AUC and \modelname{no-transfer}'s. The size of a point positively correlates with the length of input feature sequence (prediction week $k$).}
	\label{fig:transfer_scatter}
\end{figure}

The sensitivity of some of the important parameters of the \modelname{passive} and \modelname{active} methods are also investigated in Figure~\ref{fig:transfer_scatter}. We analyze how the performance of passive and active transfer is correlated to the PAD and the sample size ratio between the source and target courses. We see that the \modelname{passive} performs better than the \modelname{active} approach when transferring from a course with $\sim 6,000$ students (\fullcoursename{2}) to one with $\sim 15,000-40,000$ students (\fullcoursename{1}) (the upper right corner of Figure~\ref{fig:transfer_scatter}), and in the other cases the \modelname{active} often sightly outperforms the \modelname{passive}. We find that the best transfer performance is close to the \modelname{no-transfer} baseline when both the target to source sample size ratio and the PAD are small. \modelname{active} or \modelname{passive} transfer have on average an 8\% improvement compared with the \modelname{naive} baseline in terms of AUC scores (calculated from the last columns of Table~\ref{tab:gorup11_model_comparison} and Table~\ref{tab:gorup21_model_comparison}).

\subsection{Experience-Based Transfer Models}
\begin{figure}[t]
	\centerfloat
	\begin{subfigure}[b]{0.80\linewidth}
		\centerfloat
		\includegraphics[width=\linewidth]{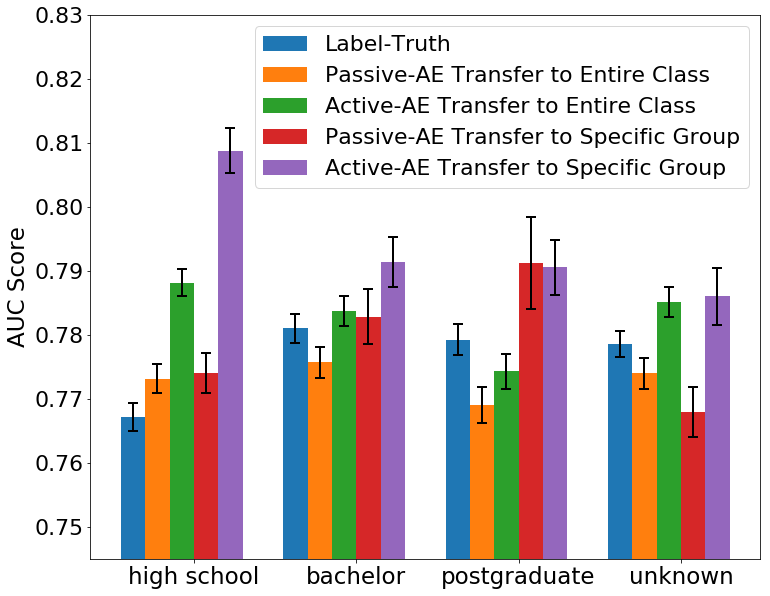}
		\caption{\label{fig:performance_education_backgrounds} Average AUCs on students with different\\educational backgrounds in \fullcoursename{2}.}
	\end{subfigure}
	\begin{subfigure}[b]{0.23\linewidth}
		\centerfloat
		\includegraphics[width=\linewidth]{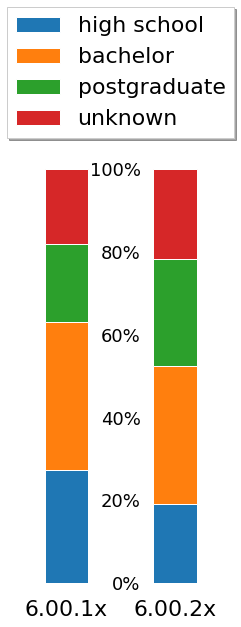}
		\caption{\label{fig:education_propotions} Class\\compositions.}
	\end{subfigure}
	\caption{(a): Averages AUC scores on students with different educational backgrounds for the \modelname{no-transfer} case, and the \modelname{passive} and \modelname{active} methods when transferring from \fullcoursename{1} to an entire \fullcoursename{2} offering or a specific group of students within that offering on all weeks on all possible source and target combinations. (b): Average percentages of students with high school, bachelor, and postgraduate degrees in \fullcoursename{1} and \fullcoursename{2}.}
	\label{fig:transfer_education}
\end{figure}

The composition of a MOOC class can be very diverse, and this is one of the differences compared to traditional education. Since \fullcoursename{2} is a more in-depth course than \fullcoursename{1}, the proportion of high school students in \fullcoursename{2} is smaller, but there are more postgraduate students (Figure~\ref{fig:education_propotions}). Training a predictive model for high school students on \fullcoursename{2} can be difficult in the sense that there are not enough samples to learn from. Transferring the knowledge from \fullcoursename{1} by the proposed methods can partially solve this problem. Equipped with the student background labels, there are two possible ways of transfer learning: transfer to the entire target class as we did in the other experiments, or consider each group of students as a target and transfer specifically to that group in an experiment. With simpler target distributions and thus easier transfer objectives, the latter approach is potent to achieve better prediction performance. We evaluate their performance when transferring from \fullcoursename{1} to \fullcoursename{2} as shown in Figure~\ref{fig:performance_education_backgrounds}. Where we see the average AUC on high school students is improved by active transfer, and is even further improved if we specifically transfer to the high school group. \modelname{passive} does not perform as well as \modelname{active}, since the source course has much more students than the target. Results showed that \modelname{active} to specific groups performed the best. This approach can also be applied to minority groups based on other demographic variables including income level, gender, age, location, ethnicity, and race.

\subsection{Examining the Embeddings}

\begin{figure}[t]
	\centerfloat
	\begin{subfigure}[b]{1.1\linewidth}
		\centerfloat
		\includegraphics[width=\linewidth]{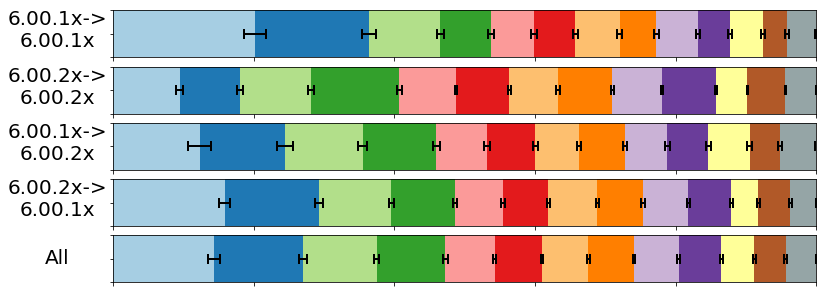}
	\end{subfigure}
	
	\begin{subfigure}[b]{0.8\linewidth}
		\centerfloat
		\includegraphics[width=\linewidth]{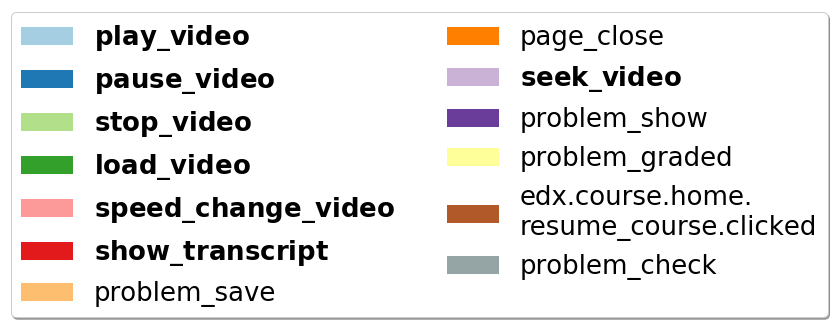}
	\end{subfigure}
	\caption{Compositions of the embeddings learned by \modelname{active} (with error bars) using the neural network implementation of PCA. The legends are ordered with decreasing importance in the average embedding (the last bar).}
	\label{fig:embedding_composition}
\end{figure}

To further examine the transfer embeddings we replaced the \textbf{LSTM-AE} used by \modelname{active} with a neural network implementation of PCA to access the embedding as a linear transformation of the raw feature space. Now we can calculate the ``weight'' (the scaling factor in the direction of a feature under the PCA transformation) of a raw feature in the learned embedding, which is an indicator of how vital the raw feature is for the transfer learning task. We calculate the average weights of the thirteen features on different groups of source and targets, as shown in Figure~\ref{fig:embedding_composition}. The average weights on all possible transfers (the last bar) show the relative importance of each raw features, and we order the legend labels accordingly. We can see that video related events (bold) occupy the first six places, which implies they are more transferable and predictive than the others.


\section{Conclusions}
\label{sec:conclusions}
In this paper, we explored the possibilities of using deep auto-encoders for transfer learning among courses in MOOCs. We proposed two methods to improve the transferability of learned embedding: the \modelname{passive} using transductive PCA to remove the variations present in the source domain but irrelevant for the target domain, and the \modelname{active} using the CORAL loss to force the alignment of the second-order statistics of the source and target embeddings. Moreover, the deep transfer models solve the domain-specific feature engineering problem and can learn compact and effective representations from the raw features, a straight-forward representation of the click-stream. We found that our transferred models consistently outperform the other transfer baselines and achieve similar performance compared with the label-truth (no transfer) models which are trained on the target course. In this sense, we believe that we have made significant progress in solving the transfer learning problem in MOOCs. With the acknowledged limitations that the models, while accurate and automated, are not transparent and do not integrate contextual human knowledge like the learning design.

We answered the following research questions:
\begin{inparaenum}
	\item Does representation learning improve model transfer? We evaluated transferability within offerings of two courses and across two courses. We found that transferring from a course with fewer students to one with more students \modelname{passive} was best versus \modelname{active} and baselines. In the opposite situation, \modelname{active} was best versus \modelname{passive} and baselines. Both passive and active methods approached the AUC level of the \modelname{no-transfer} (no transfer) baseline, which learned from target labels directly.
	\item Can representation-based learning work from a universal, basic set of MOOC activity features as input? We successfully used the same time series per student where the frequencies of a set of specific MOOC activity types are expressed per time unit as input to every experiment. It supported our transfer learning results. This suggests that, for transfer learning problems in MOOCs, in general, it is possible to eliminate costly feature engineering.
	\item Can transfer learning improve recognition of minority groups? If we group similar students and transfer learning for each group independently, does predictive performance improve? We grouped students by their highest level of education and found that transfer learning with the proposed methods can help improve the prediction performance on minority groups, and transferring specifically to a target group might achieve even higher performance. This is an example where contextual knowledge, if available, can be used to improve this methodology despite that knowledge not being explicitly integrated into the model.
	\item What are the embedded features that increase the transferability? We calculated the weights of event features in the embedding learned by \modelname{active} using PCA instead of LSTM AE and showed that video related events are more transferable and predictive than the others.
\end{inparaenum}

The contributions of this paper are:
\begin{asparaenum}
    \item We introduced two online transfer learning methods based on representation learning that improve prediction for the target course and eliminate manual feature engineering. We improve the dropout prediction AUC scores by 8\% using either method compared with the naive transfer baseline.
    \item We introduced a data organization for input to the prediction and transfer methods that requires no feature extraction. It is a time series per student where the frequencies of a set of specific MOOC activity types are expressed per time unit.
    \item Through visualization and metric analysis, we described the representation-based learning embeddings.
    \item We found that transfer learning for specific groups of students independently improved predictions, facilitating more specific learning support.
\end{asparaenum}

For future work, a direct extension of our work is to apply the transfer algorithms to different types of MOOCs and across MOOC platforms. It will also be interesting to investigate how we can effectively learn from multiple source courses, simultaneously characterizing and utilizing the sources' relationships to the target course based on course content and student population. Finally, it is possible to investigate further raw data representations that take structural course context into account, not only temporal activity.

\bibliographystyle{ACM-Reference-Format}
\bibliography{references}

\appendix
\section*{APPENDIX}
\section{Model Architectures}
\label{appendix:model_architectures}
The detailed architectures of the three neural network models used by all the transfer methods and the label-truth baseline are:
\begin{asparadesc}
    \item[LSTM Predictive Model] Input--Conv1D(16, 1)--ReLU\\--Conv1D(8, 1)--ReLU--LSTM(8)--ReLU--Flatten--FC(1)--Sigmoid\\--Output
    \item[LSTM AE] Input--Conv1D(12, 1)--LeakyRelu(0.2)--BLSTM(8)\\--LeakyRelu(0.2)--Conv1D(8, 1)--Flatten--Embedding--Reshape\\--BLSTM(6)--LeakyRelu(0.2)--Conv1D(13, 1)--Sigmoid--Output
    \item[CNN Predictive Model on AE embeddings] Input\\--Conv1D(8, 3)--ReLU--Conv1D(8, 3)--ReLU--Flatten--FC(1)\\--Sigmoid--Output
\end{asparadesc}
Where FC(n) is a fully connect layer with $n$ neurons, Conv1D(n, k) is an 1D-convolutional layer with $n$ output channels and kernel size $k$, LSTM(n) is a LSTM layer with $n$ cells, LeakyReLU($\alpha$) is a leaky-ReLU activation, and BLSTM(n) is a bi-directional LSTM layer with $n$ cells.

\section{Experimental Configurations}
\label{appendix:experimental_configurations}
We implemented the models using \textit{PyTorch} and \textit{Keras} in \textrm{Python}. For all training, the batch size is $128$, the \textit{Adam} optimizer is used with learning rate $0.001$, and the number of epochs is between $100$ and $200$. For all AEs, the bottleneck size is eight dimensions per time unit. For \modelname{no-transfer} and \modelname{no-transfer-AE}, the train-test split ratio is $4\colon1$. For \modelname{passive}, the number of output components of T-PCA is set to six per time unit. For \modelname{active}, the loss weights of the source prediction cross-entropy loss, the target autoencoding mean-squared error and the CORAL loss are $0.008$, $1$ and $1000$ respectively. For each combination of training method, model architecture, source, target, and week we only train once, since all of our results are averages with low variance, and the variance from stochastic optimization here is very small (less than 1\% according to our experiments).
\end{document}